# Fuzzy Recommendations in Marketing Campaigns

S. Podapati, L. Lundberg, L. Skold, O. Rosander, J. Sidorova

## Abstract

The population in Sweden is growing rapidly due to immigration. In this light, the issue of infrastructure upgrades to provide telecommunication services is of importance. New antennas can be installed at hot spots of user demand, which will require an investment, and/or the clientele expansion can be carried out in a planned manner to promote the exploitation of the infrastructure in the less loaded geographical zones. In this paper, we explore the second alternative. Informally speaking, the term Infrastructure-Stressing describes a user who stays in the zones of high demand, which are prone to produce service failures, if further loaded. We have studied the Infrastructure-Stressing population in the light of their correlation with geo-demographic segments. This is motivated by the fact that specific geo-demographic segments can be targeted via marketing campaigns. Fuzzy logic is applied to create an interface between big data, numeric methods for processing big data and a manager.

## 1 Introduction

In the era of big data a mapping is desired from multitudes of numeric data to a useful summary and insights expressed in a natural language yet with a mathematical precision [Zadeh, 2009]. Fuzzy logic bridges from mathematics to the way humans reason and the way the human world operates. Clearly, the "class of all real numbers which are much greater than 1," or "the class of beautiful women," or "the class of tall men," do not constitute classes or sets in the usual mathematical sense of these terms. Yet, "the fact remains that such imprecisely defined notions play an important role in human thinking, particularly in the domains of decision-making, abstraction and communication of information" [Zadeh, 1965]. According to [Meyer, Zimmerman, 2011], few works exist in business intelligence that use fuzzy logic due to certain inherent difficulties of creating such applications, and yet; despite them, such applications are possible and very useful. The difficulties are as follows. Firstly, many applications do not permit a trial and error calibration, because the results of a fuzzy model cannot easily be compared to the results of the behaviour of the real system. Secondly, the operators, membership functions, and inference methods have to properly map the counterparts of human mind, in which they are very often very context dependent. Thirdly, this is no longer a mathematical problem but predominantly a problem of psycholinguistics or similar disciplines, and unlikely this part of science is much less developed than the mathematics of fuzzy set theory. The main two types of fuzzy technology are fuzzy knowledge based systems, e.g. [Meyer, Zimmerman, 2011] and fuzzy clustering e.g. [Tettamanzi et al, 2007].

Our idea is different from the above. Fuzzy logic enables us to formulate a natural language interface between big data, numeric analytics, and a manager, hiding the compexity of data and methods. We summarize data using linguistic hedges and formulating queries in a natural language. Our specific application is targeting different user segments to fill in the spare capacity of the network in a network-friendly manner. In [Sidorova et al, 2017], the notion of *Infrastructure-Stressing* (IS) Client was proposed together with the method to reveal such clients from the customer base. Informally, IS clients use the infrastructure in a stressing manner, such as always staying in the zones of high demand, where the antennas are prone to service failures, if further loaded. Being IS is not only a function of user's qualities, but also of the infrastructure, and of the relative mobility of the rest of the population. It is not possible to directly use this knowledge in marketing campaigns, where the desired action is to avoid recruiting IS clients, at least recruiting them in disproportionally large quantities. This paper aims to make the knowledge about IS users applicable in marketing.

For marketing campaigns geodemographic segmenations (like ACORN or MOSAIC) are used, since it is known how the segments can be targeted to achieve the desired goal, as for example, the promotion of a new mobile service in certain neigbourhoods. The client's home address determines the geodemographic category. People of similar social status and lifestyle tend to live close. Compared to conventional occupational measures of social class, postcode classifications typically achieve higher levels of discrimination, whether averaged across a random basket of behaviors recorded on the Target Group Index or surveys of citizen satisfaction with the provision of local authority services. One of the reasons that segmentation systems like MOSAIC are so effective is that they are created by combining statistical averages for both census data and consumer spending data in pre-defined geographical units [Grubesic, 2004]. The postcode descriptors allow us powerful means to unravel lifestyle differences in ways that are difficult to distinguish using conventional survey research given limited sources and sample size constraints [Webber and Butler, 2007]. For example, it was demonstrated that middle-class MOSAIC categories in the UK such as 'New Urban Colonists', 'Bungalow Retirement', 'Gentrified Villages' and 'Conservative Values', whilst very similar in terms of overall social status, nonetheless register widely different public attitudes and voting intentions, show support for different kinds of charities and preferences for different media as well as different forms of consumption. Geodemographic categories correlate to diabetes propensity [Levy, 2006], school students' performance [Webber and Butler, 2007], broadband access and availability [Grubesic, 2004] and so on. Industries rely increasingly on geodemographic segmentation to classify their markets when acquiring new customers [Haenlein and Kaplan, 2009]. The localized versions of MOSAIC have been developed for a number of countries, including the USA and most of the EU countries. The main geodemographic systems are in competition with each other and the exact details of the data and methods for generating lifestyles segments are never released [Debenham et al., 2003] and, as a result, the specific variables or the derivations of these variables are unknown. To conclude, geodemographic segmentation provides a *collective view point*, where the client is seen as a representative of the population who live nearby. However, in recent research, it has been shown that the problem of resource allocation in the zones with nearly overloaded and underloaded antennas is better handled relying on *individual modelling* based on client's historical trajectories [Sagar, 2016]. The author completed a user segmentation based on clustering of user trajectories and demonstrated that network planning is more effective, if trajectory-based segments are used instead of MOSAIC segments.

Our aim is to explore the ways to connect the individual trajectory-based view of IS customers and the geo-demographic view in order to devise analytics capable to complete the efficient analysis based on the individual view point and yet be useful in marketing campaigns in which geodemographic groups are targeted. As a practical conclusion, we have compiled a ranked list of the segments according to their propensity to contain IS clients (expressed as a fuzzy notion) and crafted two queries:

1. Which segments are more or less devoid of IS clients? (attract them, while the infrastructure is still rather underloaded)
2. Which segment is highly devoid of IS clients? (when the custormer base becomes mature and the infrastructure becomes increasingly loaded).

The contributions of this paper are as follows. Firstly, we have studied the correlation between IS users and the MOSAIC segments. For different contexts, we have completed candidate rankings of geodemographic segments, and, given an absense of other preferences, the top-tier segments are preferable. Which ranking out of several candidate ones is taken depends on the hedge (degree) calculated for the intensity of infrastructure exploitation. Secondly, the verification/simulation of the resulting fuzzy recommendations guarantees the absence of false negatives, such as, concluding that certain segments can be hired from, but in fact that would lead to a service failure at some place in the network.

The rest of the paper is organised as follows. Section 2 describes the data set. In Section 3 the proposed methodology is explained. In Section 4, experiments are reported, and finally the conclusions are drawn and discussion is held in Section 5.

## 2 Data Set

The study has been conducted on anonymized geospatial and geo-demographic data provided by a Scandinavian telecommunication operator. The data consist of CDRs (Call Detail Records) containing historical location data and calls made during one week in a midsized region in Sweden with more than one thousand radio cells. Several cells can be located on the same antenna. The cell density varies in different areas and is higher in city centers, compared to rural areas. The locations of 27010 clients are registered together with which cell serves the client. The location is registered every five minutes. In the periods when the client does not generate any traffic, she does not make any impact on the infrastructure and such periods of inactivity are not included in the resource allocation analysis. Every client in the database is labeled with her MOSAIC segment. The fields of the database used in this study are:
- the cells IDs with the information about which a user it served at different time points,
- the location coordinates of the cells,
- the time stamps of every event (5 minute resolution),
- the MOSAIC geodemographic segment for each client, and
- the Telenor geodemographic segment for each client.

There are 14 MOSAIC segments present in the database; for their detailed description the reader is refferred to [InsightOne]. The six in-house Telenor segments were developed by Telenor in collaboration with InsightOne, and, to our best knowledge, though not conceptually different from MOSAIC, they are especially crafted for telecommunication businesses.

## 3 A Link between IS and Geo-demographic Segments

### 3.1 Notation and Definitions

**Definition** (in the style of [Zadeh, 1965]). A fuzzy set $A$ in $X$ is characterized by a membership function $f_A(x)$, which associates with each point in $X$ a real number in the interval $[0, 1]$, with the value of $f_A(x)$ at x representing the "grade of membership" of $x$ in $A$. For the opposite quality: $f_{notA}(x) = 1 - f_A(x)$.

Fuzzy membership scores reflect the varying degree to which different cases belong to a set:

- Under the six value fuzzy set, there are six tiers of membership $1$: fully in, $0.9$: mostly but not fully in, $0.6$: more or less in, $0.4$: more or less out, $0.1$: mostly but not fully out, $0$: fully out.
- Thus, fuzzy sets combine qualitative and quantitative assessment: 1 and 0 are qualitative assignments ("fully in" and "fully out", respectively); values between 0 and 1 indicate parcial membership. The 0.5 score is also qualitatively anchored, for it indicates the point of maximum ambiguity (fuzziness) in the assessment of whether a case is more "in" or "out" of a set.

For a comprehensive guide of good practices in fuzzy logic analysis in social sciences the reader is refferred to, for example, [Ragin, 2009].

**Interpretation:**

- *Rather* will be added to a quality $A$, if the square root of its membership function $f_A(x)^{1/2}$ is close to $1$.
- *Very* will be added to a quality $A$, if the square of its membership function $f_A(x)^2$ is close to $1$.
- *Extremely* will be added to a quality $A$, if $f_A(x)^3$ is close to $1$.

The interpretation follows from the application of the hedge operator, which adds the quantifiers such as *very, rather, extremely*, to the membership function, for example $f_{veryA}(x) = f_A(x)^2$ [Zadeh, 1972]. Then, given the new membership function, the same principle applies: the closer to 1, the higher is the degree of membership. Inside a tier, the hedged membership functions obey an inclusion relation: *extremely $f \subset$ very $f \subset$ rather $f$*.

### 3.2 Query Formulation

As mentioned above, within the same geo-demographic segment, the clients differ with respect to being IS or not. When the infrastructure is not overloaded, that is, the recent historical load is still significantly smaller than the capacity, then virtually any client is welcome. The following two queries are formulated reflecting the desire to apply context-dependent strategies. As the infrastructure becomes more loaded, the operator wants to be more discriminative with respect to the degree of the IS/IF quality. In its turn, "loaded" for an antenna is naturally formulated as a fuzzy variable:

$f_{loaded}(\text{antenna } j) = \max_{all\ t} \{load(j,t)\ capacity(\text{antenna } j)^{-1}\}$.

The $f_{loaded}(\text{antenna } j)$ is calculated in man units. The load in the analyzed zone is set to the maximum peak of demand registered:

$f_{loaded}(\text{infrastructure}) = \max_{all\ antennas\ j} \{f_{loaded}(\text{antenna } j)\}$.

**Queries:**

- Which segments to target, provided that *rather* IF are acceptable clientele?
- Which segments to target, provided that only *very* IF are wanted?

Depending on the load, different rankings of segments become available. If initially some segments were in the same tier, for example, very IF segments, some of them fall out of the tier, as the hedge operator is applied and the value of the membership function is squared (for extremely IF). The context, when to apply Query 1 or 2, becomes clarified comparing the network load (measured as network peak load) to network capacity. The method to obtain fuzzy heuristics is summarized to the sequence of the following steps, depicted as a flow chart in Figure 1, and formalized as Algorithm 1.

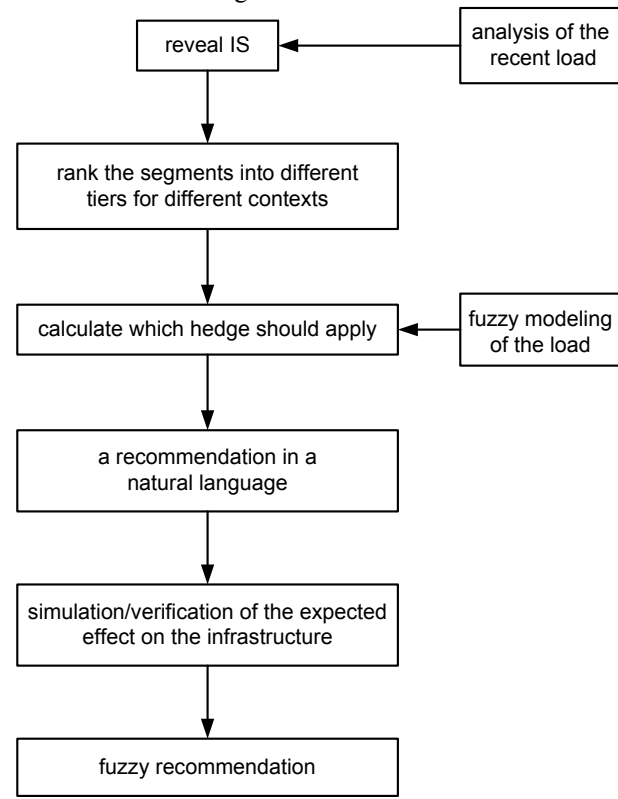

**Figure 1:** The flow chart for the calculation of fuzzy recommendation for a marketing campaign.

**Step 1:** The IS clients in the customer base are revealed with the method [Sidorova et al, 2017] (the algorithm is reproduced as the function *reveal_IS* clients in the Algorithm 1), and each client is labeled with the IS/notIS descriptor.

**Step 2:** The propensity of a segment to contain IS clients is defined as the frequency of IS clients among its members and it is calculated from the data:

$$f_{IS}(segment_i) = frequency_{IS}(segment_i).$$

Infrastructure-Friendly (IF) is set to:

$$f_{IF}(segment_i) = 1 - f_{IS}(segment_i).$$

**Step 3:** The ranking of segments is carried out with respect to their IF quality: for all segments $i$, $f_{rather\ IF}(segment_i)$, $f_{very\ IF}(segment_i)$, and $f_{extremely\ IF}(segment_i)$. Within a context, the segments fall into the different tiers (corresponding to one of the fuzzy values): "fully in", "mostly but not fully in", "more or less in", and so on.

**Step 4:** Locally for the region under analysis, the infrastructure is assessed as *loaded*, *very loaded*, and *extremely loaded*, in order to map the context into a corresponding hedge. Among several candidate rankings, the one for a corresponding hedge is selected (as a leap of faith further verified in the next section).

**Algorithm 1: computation of the fuzzy recommendation heuristic.**

*Variables:*
- clientSet: set of with IDs of clients;
- *I:* the set with geo-demographic segments $\{segment_1, …, segment_k\}$;
- *D:* the mobility data for a region that for each user contain client's ID, client's geo-demographic segment, time stamps when the client generated traffic, and which antenna served the client.
- $S_i$: the number of subscribers that belong to a geo-demographic segment $i$;
- $\Sigma_{all\ i}\ S_{i,t,j}$: the footprint, i.e. the number of subscribers that belong to a geo-demographic segment $i$, at time moment $t$, who are registered with a particular cell $j$;
- $C_j$: the capacity of cell $j$ in terms of how many persons it can safely handle simultaneously;
- *x*: the vector with the scaling coefficients for the geo-demographic segments or other groups such as IS clients;
- $x_{IS}$: the coeffcient for the IS segment from the vector *x*;
- $N_{t,j}$ = number of users at cell $j$ at time $t$;.

**Input:** data set D: <$user_{ID}$, time stamp t, cell j>.

```
reveal_ISclients;
for i in I{
  ratherIF[i] = false
  veryIF[i] = false
  extremelyIF[i] = false
  degreeIS = frequency(userID_IS,I)
  degreeIF = 1- degreeIS
  if (degreeIF^{1/2} ≥ 0.9) then ratherIF[i]=true
  if (degreeIF^2 ≥ 0.9) then veryIF[i]=true
  if (degreeIF^3 ≥ 0.9) then extremelyIF[i]=true

}

function reveal_ISclients{
```

```
[I. Characterize each user with respect to her
relative mobility.]
  for each user_ID {
    trajectory_ID = cell_{t1}, …, cell_{t2016};
    relativeTrajectory_ID = N_{t1,j}, …, N_{t2016,j};
    sortedTrajectory_ID =
        sort_{decreas_or.}(relativeTrajectory_ID);
    topHotSpots_ID = Σ_{k=1..100(5%)}sortedTrajectory_ID[k];
    userTopHotSpots = <user_ID, topHotSpots_ID>
  }
  rankedUserList = sort_{decreasing_or}(rankedUserList)

  [II. Initialization.]

      x_{stressing} = 0;
      setStressingUsers = ∅.

[III. Reveal the infrastructure-stressing
clients.]

  While (x_{stressing} = 0) do {
     tentativeStressingUsers =
     head_{1%}(rankedUserList);
     setFriendlyUsers = bottom_{1%}(rankedUserList);
     otherUsers = rankedUserList –
     tentativeSetStressingUsers -
     setFriendlyUsers;

     [Confirm the tentative labeling via
     combinatorial optimization.]
     I = {stressing, medium, friendly};
       {x_{stressing}, x_{medium}, x_{friendly}} =
     combinatorial_optimization(I,D);

     IF (x_{stressing} = 0), THEN {
     tentativeSetStressingUsers =
     tentativeStressingUsers< userID >^1;
     setStressingUsers = setStressingUsers +
     tentativeSetStressingUsers<UserID>

     D = D – D_{stressing}
  } [end of while]

  for id in <userIDs> do {
     if (id ∈ setStressingUsers) then
     label(id,"IS")
     else label(id,"notIS")
  } [end loop on id in <userIDs>]
} [end reveal_ISclients]

function combinatorial_optimization(I,D){
      solve
      Maximize Σ_{i={IF,other,IS}} S_i x_i,
         subject to:

         for all j,t, Σ_{i={IF,other,IS}} S_{i,t,j} x_i ≤ C_j

} returns {x_IF, x_other, x_IS}.
```

**Output:** array ratherIF[], veryIF[], extremelyIF[].

### 3.3 Query Simulation

In the above, when deciding which context should be applied, we relied on an intuitive rule: If the load is

---
[1] field userID from tentativeStressingUsers

<hedge $X^2$> big, then <hedge X> IS segments are suitable to hire clients from. It does not necessary hold, since the calibration of fuzzy functions depend on the expected outcome of the campaign and the consequent effect on the infrastructure. For example, the campaign can attract 300 new clients or 1500 new clients. To avoid false negatives, the fuzzy heuristic is subjected to a validation procedure, which simulates the impact of the expected result on the infrastructure.

*I. It throws a warning, if some antenna is overloaded.* That is, if expected footprint by the segment violates a restriction for some segment $i$, some antenna $j$, some time moment $t$:

$$\alpha S_{i,j,t} \leq C_j,$$

where $\alpha$ is a scaling coefficient,
$\alpha$ = expected number of new clients × (total number of clients)$^{-1}$. This is a justifiable approximation, because of the consensus in the literature is that there is a high predictability in user trajectories within different segments, e.g. [Song et al, 2010], [Lu et al, 2013].

*II. It recalculates the hedge for "being loaded".*

## 4 Experiment

**1. Reveal the IS clients.** Applying the algorithm to reveal IS clients from, we have added a field to data set with the label IS or Not IS as a descriptor for each client.

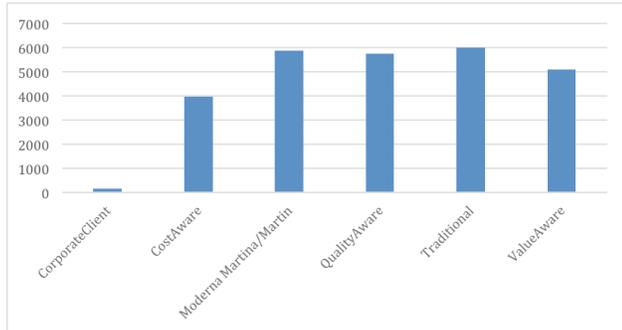

**Figure 2:** The distribution of users across Telenor segmetns in the region.

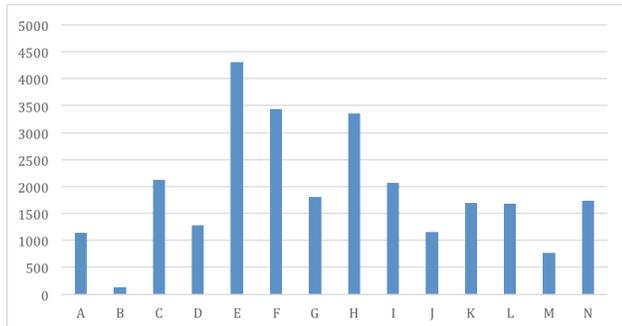

**Figure 3:** The distribution of users across MOSAIC segmetns in the region.

---
[2] for example, *very*.

**2. Calculate degree of infrastructure-friendliness for each segment.** The charts with the number of customers in each MOSAIC and Telenor segments in the geographic region are represented in Figures 2 and 3. In the whole customer base, *7%* of subscribers were revealed to be IS [Sidorova et al, 2017]. We have obtained the distribution of the IS clients within the MOSAIC and Telenor segments and depicted them in figures 4 and 5, respectively. The degree of the infrastructure-friendliness is reported in Table 1 and 2, for MOSAIC and Telenor segments, respectively.

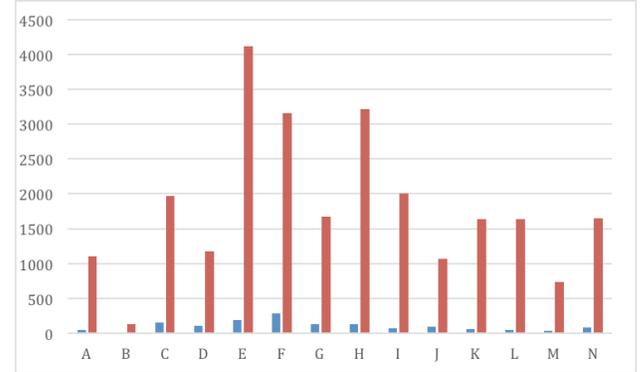

**Figure 4:** The percent of IS clients in different MOSAIC categories.

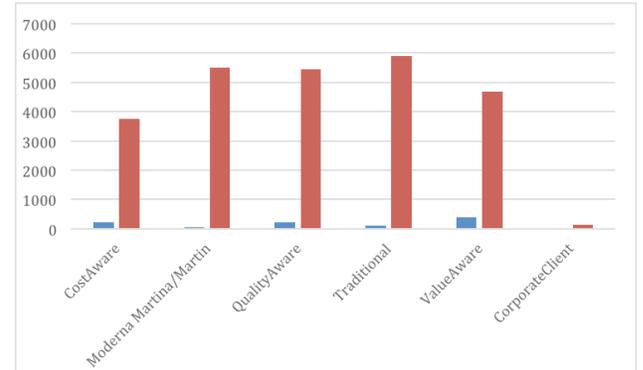

**Figure 5:** The percent of IS clients in different Telenor segments.

**II. Reasoning behind the queries.** Tables 1 and 2 simulate the reasoning behind the query results for different contexts (codified with a hedge) for the MOSAIC and Telenor segments, respectively. Each of the *14* MOSAIC classes qualifies as *rather IF,* which are those with $f_{IF}(i)^{1/2} > 0.9$. Once the customer base becomes larger and the spare capacity diminishes, only *very IF* will be wanted, which are those with $f_{IF}(i)^2 > 0.9$. Out of those 9 segments qualify as very IF and five segments qualify as extremely IF $(f_{IF}(i)^3 > 0.9)$. The customer population was subjected to the same analysis with respect to Telenor segmentation. As follows from Table 2, each of the six

Telenor segments is rather friendly, and there are four and three very and extremely IF segments, respectively.

| segment | $f_{IF}(i)$ | $f_{IF}(i)^{1/2}$ | rather IF? | $f_{IF}(i)^2$ | very IF? | $f_{IF}(i)^3$ | extremely IF? |
|---|---|---|---|---|---|---|---|
| A | 0.96 | 0.97 | yes | 0.92 | yes | 0.88 | no |
| B | 0.98 | 0.98 | yes | 0.96 | yes | 0.94 | yes |
| C | 0.93 | 0.96 | yes | 0.86 | no | 0.79 | no |
| D | 0.92 | 0.95 | yes | 0.84 | no | 0.77 | no |
| E | 0.96 | 0.97 | yes | 0.92 | yes | 0.88 | no |
| F | 0.92 | 0.95 | yes | 0.86 | no | 0.79 | no |
| G | 0.93 | 0.96 | yes | 0.86 | no | 0.79 | no |
| H | 0.96 | 0.97 | yes | 0.92 | yes | 0.88 | no |
| I | 0.97 | 0.98 | yes | 0.94 | yes | 0.91 | yes |
| J | 0.92 | 0.95 | yes | 0.86 | no | 0.79 | no |
| K | 0.97 | 0.98 | yes | 0.94 | yes | 0.91 | yes |
| L | 0.98 | 0.98 | yes | 0.96 | yes | 0.94 | yes |
| M | 0.96 | 0.97 | yes | 0.92 | yes | 0.88 | no |
| N | 0.95 | 0.97 | yes | 0.9 | yes | 0.85 | no |

**Table 1:** The reasoning behind the query results for the MOSAIC segments.

| segment | $f_{IF}(i)$ | $f_{IF}(i)^{1/2}$ | rather IF? | $f_{IF}(i)^2$ | very IF? | $f_{IF}(i)^3$ | extremely IF? |
|---|---|---|---|---|---|---|---|
| CA | 0.94 | 0.97 | yes | 0.88 | no | 0.82 | no |
| MM | 0.99 | 0.89 | yes | 0.98 | yes | 0.97 | yes |
| QA | 0.96 | 0.92 | yes | 0.92 | yes | 0.88 | no |
| T | 0.98 | 0.87 | yes | 0.96 | yes | 0.94 | yes |
| CC | 0.92 | 0.8 | yes | 0.86 | no | 0.79 | no |
| VA | 0.97 | 0.91 | yes | 0.94 | yes | 0.91 | yes |

**Table 2:** The reasoning behind the query results for the Telenor segments.

## 6 Results

When it comes to designing strategies of accomodating many more clients, being IS-prone for a segment is an important quality. We have studied the correlation between IS users and the MOSAIC segments, motivated by the fact that we can target the MOSAIC segments in marketing campaigns. For different contexts, we have completed candidate rankings of geodemographic segments, and, given an absense of other preferences, the top-tier segments are preferable. Which ranking out of several candidate ones is taken depends on the hedge calculated for the intensiveness of infrastructure exploitation. The verification/simulation guarantees no false negatives, such as saying that certain segments are safe to hire from, but in fact that would lead to a service failure at some place in the network.

## Acknowledgments